\documentclass[letterpaper]{article} 
\usepackage{aaai25}  
\usepackage{times}  
\usepackage{helvet}  
\usepackage{courier}  
\usepackage[hyphens]{url}  
\usepackage{graphicx} 
\usepackage{soul}
\usepackage{multirow}
\usepackage{booktabs}
\usepackage{amssymb}
\usepackage{amsmath}
\usepackage{subcaption}
\usepackage{bbding} 
\usepackage{natbib}  
\usepackage{caption} 
\usepackage{algorithm}
\usepackage{algorithmic}

\usepackage{newfloat}
\usepackage{listings}
\DeclareCaptionStyle{ruled}{labelfont=normalfont,labelsep=colon,strut=off} 
\floatstyle{ruled}
\newfloat{listing}{tb}{lst}{}
\floatname{listing}{Listing}
\usepackage{xcolor}
\definecolor{sportscolor}{rgb}{0, 0.6, 0.2}
\definecolor{mycolor}{RGB}{255, 0, 0}
\definecolor{mycolor}{RGB}{255, 0, 0}

\title{Controlling Large Language Models Through Concept Activation Vectors}
\author{
    Hanyu Zhang\textsuperscript{\rm 1,2,3}
    , Xiting Wang\textsuperscript{\rm 4}\thanks{Corresponding authors.}, Chengao Li\textsuperscript{\rm 1,2,3}, Xiang Ao\textsuperscript{\rm 1,2,3}, Qing He\textsuperscript{\rm 1,2,3,*}
}

\affiliations{
    \textsuperscript{\rm 1}Key Lab of Intelligent Information Processing of Chinese Academy of Sciences (CAS), \\
Institute of Computing Technology, CAS, Beijing 100190, China\\
    \textsuperscript{\rm 2}Key Lab of AI Safety of Chinese Academy of Sciences (CAS),  Beijing 100190, China\\
\textsuperscript{\rm 3}University of Chinese Academy of Sciences, CAS, Beijing 100049, China \\
\textsuperscript{\rm 4} Gaoling School of Artificial Intelligence, Renmin University of China, Beijing, China \\
\{zhanghanyu20z, lichengao22z, aoxiang, heqing\}@ict.ac.cn, xitingwang@ruc.edu.cn\\
}

\usepackage{bibentry}

\begin{document}

\maketitle

\begin{abstract}
As large language models (LLMs) are widely deployed across various domains, the ability to control their generated outputs has become more critical. This control involves aligning LLMs outputs with human values and ethical principles or customizing LLMs on specific topics or styles for individual users. Existing controlled generation methods either require significant computational resources and extensive trial-and-error or provide coarse-grained control.  In this paper, we propose Generation with Concept Activation Vector (GCAV), a lightweight model control framework that ensures accurate control without requiring resource-extensive fine-tuning. Specifically, GCAV first trains a concept activation vector for specified concepts to be controlled, such as toxicity. During inference, GCAV steers the concept vector in LLMs, for example, by removing the toxicity concept vector from the activation layers. Control experiments from different perspectives, including toxicity reduction, sentiment control, linguistic style, and topic control, demonstrate that our framework achieves state-of-the-art performance with granular control, allowing for fine-grained adjustments of both the steering layers and the steering magnitudes for individual samples.

\end{abstract}

%

\section{Introduction}

Large Language Models (LLMs)~\cite{NEURIPS2020_1457c0d6, chowdhery2023palm, touvron2023llama} have shown remarkable performance in a variety of tasks, including question answering~\cite{shi2024replug, wei2022emergent}, symbolic reasoning~\cite{hu2023chatdb, pan2023logic}, and code generation~\cite{roziere2023code}. These models are typically pre-trained on vast and diverse datasets sourced from the internet, encompassing a broad spectrum of human knowledge and interactions~\cite{peters-etal-2018-deep, devlin2018bert}. As a result, LLMs have become foundational to many Natural Language Processing (NLP) applications. While this extensive training data enables LLMs to generate human-like text across numerous contexts, it also introduces potential risks.  The data can contain unsafe content such as toxicity~\cite{gehman-etal-2020-realtoxicityprompts}, bias~\cite{gallegos2024bias}, misinformation~\cite{cao2024can, chen2023combating}, and other undesirable elements, leading to problematic LLM outputs like toxicity or hallucination~\cite{bang2023multitask}. Therefore, controlled LLM generation is particularly crucial.

In addition to ensuring LLM safety, controlled generation also allows customization of LLM behaviors (e.g., output topics and styles), which becomes increasingly important in different applications \cite{dekoninckcontrolled}. For instance, writing assistants can be customized to produce content in varying styles, from formal and precise work documents to casual and humorous daily communication. Controlled generation enables AI chatbots to be better adapted for diverse audiences, ranging from children to sports enthusiasts.

A common technique for controlled text generation is prompting engineering~\cite{sahoo2024systematic},  which is easy to implement. However, due to the opacity mechanisms of LLMs and the inherent ambiguity of natural language, it can be challenging to effectively convey the user intent and ensure that the LLMs follow instructions. 
For example, prompting an LLM with instructions like `Don't generate monkeys' can paradoxically increase the likelihood of the model referencing  `monkeys', contrary to the original intention~\cite{jang2023can}. Moreover, prompt engineering can be rigid, resulting in repetitive or limited responses and lacking the flexibility to adjust the level of control~\cite{li2024long}.
Another approach is parameter fine-tuning~\cite{ schulman2017proximal, ouyang2022training}, which demands substantial computational resources and is impractical for many users or real-time applications. Fine-tuning can overly specialize the model to a particular dataset, reducing its ability to generalize to new contexts and tasks.
Guided decoding is another approach~\cite{Dathathri2020Plug, yang2021fudge}, which manipulates the probability distribution during text generation. While this approach can enhance the variety of generated text, direct intervention in the decoding process can impact output fluency (see results in ~\ref{tab: toxicity 7b main results}). Additionally, the interpretability of these methods remains a significant concern~\cite{zhong2023air}.

In this paper, we introduce a method for controlled LLM generation by modifying intermediate activation vectors during inference, a technique referred to as activation engineering~\cite{turner2023activation}.
Recent works have shown that certain directions in the activation space are associated with semantic attributes \cite{luo2024pace}. However, a key challenge remains: how to accurately calculate the direction of a concept and then precisely steer the direction vector for each input sample while maintaining fluency. To address this, we propose a novel framework called \textit{Generation with Concept Activation Vectors (GCAV)}, inspired by the explainable approach of Concept Activation Vectors used in model decision interpretation~\cite{kim2018interpretability}. GCAV framework trains a concept activation vector for a specified concept, such as toxicity, and then steers the vector to LLMs to control this specific concept, for example, by removing the concept toxicity. 
Specifically, we construct a small set of contrastive prompts (e.g., 100 pairs) to guide the LLM in generating content either with or without the target concept, then collect the corresponding activation vectors for classification. During inference, the concept activation vector is applied to the selected layers with a calculated steer strength. This approach enables granular control over LLMs generation, ensuring the outputs align with the intended properties.

Our main contributions are summarized as follows:
\begin{itemize}
    \item We propose a lightweight framework for controlled LLM generation that does not require fine-tuning the model. It could achieve granular control by calculating a steering weight for each input.
    \item The GCAV framework can also control multiple concepts simultaneously, allowing for the addition or removal of various attributes as needed.
    \item Experiments demonstrate that our GCAV framework has excellent control capabilities in multiple aspects, including toxicity reduction, sentiment control, topic control, and linguistic style control.
\end{itemize}

\section{Related Work}

\paragraph{Controlled Text Generation.} ~ Controlled text generation (CTG)~\cite{zhang2023survey} aims to control the output of LLMs to meet specific criteria, such as safety standards, emotional tones, or thematic requirements. Early approaches primarily leverage prompt engineering~\cite{brown2020language} as a straightforward method to guide the generation process~\cite{li2021prefix, wei2022chain, yao2024tree}. Prompting-based CTG is intuitive and can effectively align generated contents with broad attributes~\cite{yang2022tailor}. However, the inherent ambiguity of natural language makes it difficult to express specific attributes accurately through prompts. Additionally, LLMs sometimes struggle to rigorously follow instructions~\cite{jang2023can}. 
Subsequent advancements focus on combining Supervised Fine-Tuning (SFT) with Reinforcement Learning from Human Feedback (RLHF)~\cite{ schulman2017proximal, ouyang2022training}. This paradigm involves directly modifying the model parameters to refine the model behavior. However, this approach relies on highly specific training data and specialized fine-tuning of the base model, which limits its adaptability across different models. 
An alternative strategy involves adjusting token probabilities during the decoding phase, allowing control over generations without altering the model parameters~\cite{pei2023preadd, dekoninckcontrolled}. These methods can be applied to various LLMs. ~\citet{Dathathri2020Plug}, ~\citet{yang2021fudge} use small models to guide the decoding process of LLMs, imposing constraints on the generated text to achieve specific goals. However, such external control can sometimes degrade the naturalness and fluency of the output, affecting overall text quality~\cite{zhong2023air}.

\paragraph{Activation Engineering.} ~ Activation engineering involves manipulating the internal activations of LLMs to influence their behavior and outputs in tasks such as decision-making~\cite{li2023emergent, nanda2023emergent} and sentiment analysis~\cite{tigges2023linear}. In the context of CTG, recent studies have demonstrated that certain directions in the activation space of LLMs are associated with semantic attributes~\cite{turner2023activation, luo2024pace}. By adjusting these neural activations, it is possible to achieve fine-grained control over the generated content to ensure alignment with desired attributes~\cite{zou2023representation}. Compared to traditional approaches like prompt engineering or fine-tuning, activation engineering provides a more direct and interpretable method for controlling model behaviors and outputs. However, a key challenge in activation engineering for CTG is to decide the correct activation directions and precisely control these activation manipulations.

\paragraph{Concept Activation Vector.} ~ Concept Activation Vectors (CAVs), first introduced by ~\citet{kim2018interpretability}, provide a method for quantifying a model's sensitivity to specific human-interpretable concepts by leveraging the directional derivatives of its activations.  Although initially developed for computer vision applications, CAVs have since been widely adopted in tasks involving LLMs. \citet{xu2024uncovering} used CAVs to interpret the safety mechanisms of LLMs. \citet{liucontext} and  ~\citet{todd2024function} use similar semantic vectors, such as in-context vectors (ICVs) and function vectors (FVs), to shift the latent states of LLMs during in-context learning. 

\section{GCAV Framework}

We begin by defining the problem formulation. Consider an LLM with $L$  layers. Given an input $x$, the LLM produces a sequence of activation vectors $\{\boldsymbol{e}^{(1)}, \dots, \boldsymbol{e}^{(L)}\}$ after each layer.
For a concept of interest, our objective is to modify these activation vectors $\boldsymbol{e}^{(i)}$ to new vectors $\phi_i(\boldsymbol{e}^{(i)})$, which are then fed into the subsequent layers of the model. 
This modification process aims to control the final LLM response, ensuring it adheres to the desired properties related to the specified concept.

\begin{figure*}[h]
    \centering
    \includegraphics[width=0.8\textwidth]{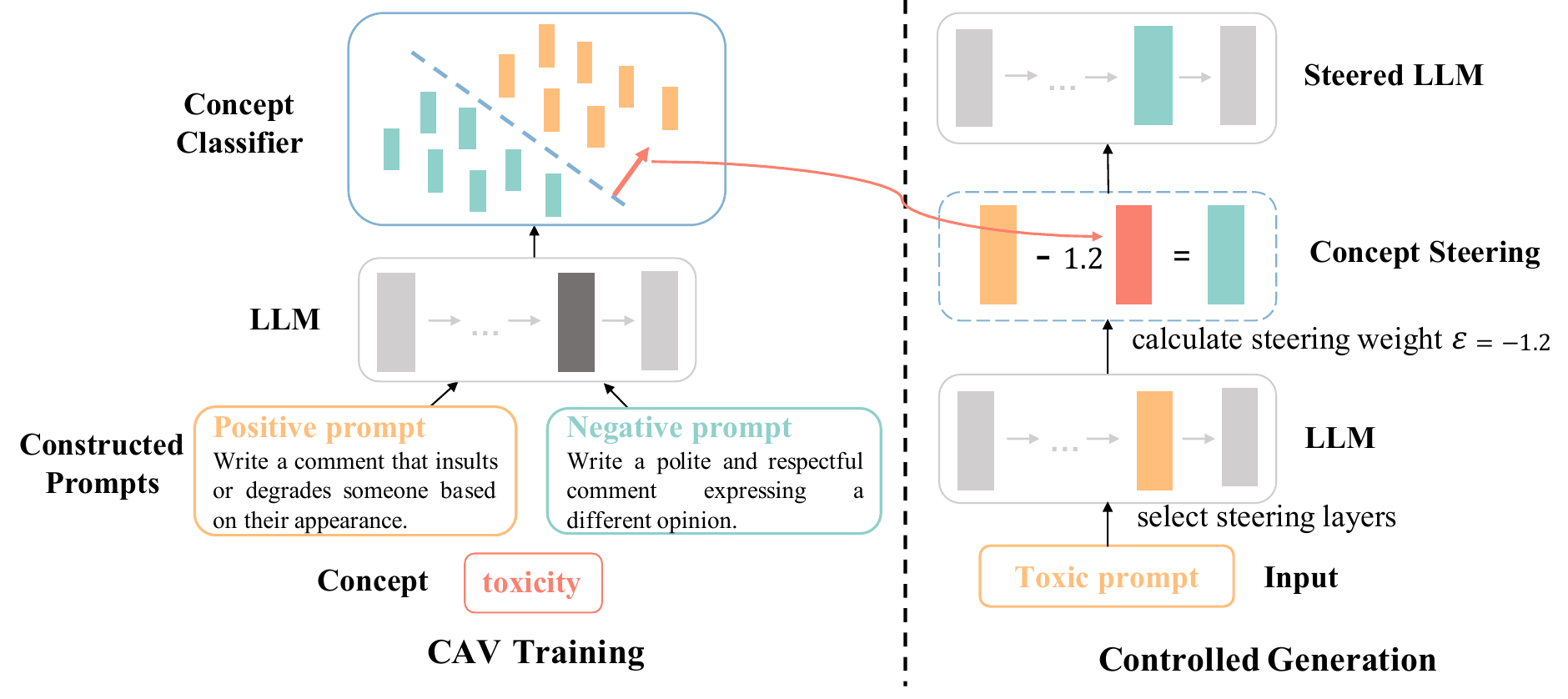}
    \caption{\textbf{CAV Training (left)}: For a given concept, such as toxicity, we construct contrastive prompts that guide the LLM to generate toxic and safe outputs. Next, we collect the activation vectors after each LLM layer and use a classifier to distinguish these two classes of activation vectors. The normal direction vector of the classifier represents the learned Concept Activation Vector (CAV).  
    \textbf{Controlled Generation (right)}: For any toxic input, we select specific LLM layers and steer the learned CAV to these layers with a calculated strength, thereby controlling the LLM generation.}
    \label{fig: GCAV}
\end{figure*}

The GCAV framework is illustrated in Figure~\ref{fig: GCAV}. First, we collect contrastive data related to a given concept and then use them to learn a corresponding concept vector. This vector is subsequently steered into the LLM with calculated weights, enabling us to control generation concerning the specified concept. The following sections will introduce the details of this process.

\subsection{CAV Training}

Our method is inspired by the Concept Activation Vector (CAV)~\cite{kim2018interpretability}, which is an explainable method to interpret how neural network internal representations work in model decisions.
Given a concept, such as toxicity, and an activation layer $l$, we train a classifier to model whether the activation vector $\boldsymbol{e}^{(L)}$ will cause the LLMs to generate outputs containing the concept (toxicity). From this classifier, we obtain the concept activation vector $v^{(L)}$ for layer $l$, which represents the specific concept.

Specifically, we first collect data to train the activation vector classifier. For a given concept, such as toxicity, the core idea is to create contrastive data pairs centered around this concept. LLMs are prompted to generate both toxic and non-toxic content using toxicity and non-toxicity prefixes. Alternatively, LLMs can be prompted with questions related to a specific concept, such as `child,' and a contrasting concept, such as `adult.' We then collect the activation vectors at each layer. The activation vectors associated with the target concept serve as positive training samples, while those related to the other concept are used as negative samples. We refer to this approach as \textbf{GCAV-Input}, as the classifier is trained on data generated from different classes of input prompts. 
To further refine this, we filter these two classes of prompts to ensure that the LLMs' responses are indeed concept-related or concept-unrelated. We then train the activation vector classifier accordingly, a method which we refer to as \textbf{GCAV-Output}.

Then, we use logistic regression as the classifier for our approach. The probability that given the activation vector $\boldsymbol{e}^{(l)}$, the output $O$ is related to concept $d$ is:
\begin{equation}
P_d^{(l)}(\boldsymbol{e}^{(l)})=\operatorname{sigmoid}\left(\boldsymbol{w}_d^{(l)\top} \boldsymbol{e}^{(l)}+b_d^{(l)}\right)
\end{equation}
where $\boldsymbol{w}_d^{(l)\top} $ and $b_d^{(l)}$ are the classifier parameters for concept $d$ and layer $l$. 

The concept activation vector is defined as follows:
\begin{equation}
\boldsymbol{v}^{(l)}=\frac{\boldsymbol{w}^{(l)}}{\|\boldsymbol{w}^{(l)}\|}
\end{equation}
This vector represents the classifier's normal direction, which is perpendicular to the decision boundary. It points directly toward the region associated with the positive class, indicating the presence of a specific concept, such as toxicity. Therefore, we can amplify the concept by adding the vector or remove the concept by subtracting the vector.

\begin{table*}[]
\centering
\resizebox{0.65\textwidth}{!}{
\begin{tabular}{lcccccc}
\toprule

                & \multicolumn{3}{c}{toxicity\_ toxic}       & \multicolumn{3}{c}{toxicity\_ random}        \\
                \cmidrule(r){2-4} \cmidrule(r){5-7}
                & toxicity $\downarrow$ & perplexity$\downarrow$ & fluency $\downarrow$ & toxicity$\downarrow$ & perplexity$\downarrow$ & fluency$\downarrow$ \\
\midrule
BASE            & 0.1807   & 13.7060    & 74.8782           & 0.0956   & 19.2312    & 79.0786             \\
POSPROMPT       & 0.1913   & 59.9855    & 91.2385           & 0.1008   & 18.3209    & 93.4485             \\
Arithmetic      & 0.1625   & \textbf{6.8436}     & 78.5721           & 0.0816   & \underline{7.3447}     & \underline{64.4872}             \\
ActAdd          & 0.1620   & 34.0770    & 100.3365          & 0.0852   & 12.6114    & 73.0775             \\
\midrule
GCAV -   Input  & \underline{0.1231}   & \underline{8.1805}     & \textbf{59.3151 }          & \underline{0.0666}   & 9.4698     & 67.5561             \\
GCAV -   Output & \textbf{0.0879}   & 21.2889    & \underline{71.7866}          & \textbf{0.0622}   & \textbf{6.0804}    & \textbf{50.2725}   \\          
\bottomrule
\end{tabular}
}
\caption{Toxicity reduction results on Llama-2-7b-chat.}
\label{tab: toxicity 7bchat main results} 
\end{table*}

\subsection{Controlled Generation}

In the LLM generation period, we employ vector addition intervention by adding or subtracting a concept direction from the latent vector $\boldsymbol{e}^{(l)}$. For instance, to remove an undesirable concept, toxicity, the intervention is expressed as: 
\begin{equation}
\boldsymbol{e}^{\prime} = \boldsymbol{e} + \epsilon \cdot \boldsymbol{v}_{\text {toxicity }}
\end{equation}
where $\boldsymbol{v}_{\text {toxicity }}$ represents the concept activation vector from the concept classifier, and $\epsilon$ is the steering strength. Here, we omit the superscript about the number of layers for simplicity of expression.

Unlike previous works that directly fix the $\epsilon$, we calculate the optimal steering strength $\epsilon$ by solving an optimization problem.
Specifically, to amplify the concept, we ensure that the probability of responses containing the concept,  given the concept vector $\boldsymbol{v}_d$, is greater than $p_d$:
\begin{equation}
\underset{\epsilon}{\arg \min }|\epsilon|, \quad \text { s.t. } P_d(\boldsymbol{e}+\epsilon \cdot \boldsymbol{v}_d) \geq p_d
\label{equ: steer plus}
\end{equation}
Conversely, when removing the concept, the probability should be less than $p_0$:
\begin{equation}
\underset{\epsilon}{\arg \min }|\epsilon|, \quad \text { s.t. } P_d(\boldsymbol{e}+\epsilon \cdot \boldsymbol{v}_d) \leq p_d
\label{equ: steer minus}
\end{equation}

The optimization problem for equation(\ref{equ: steer plus}) has a closed-form solution:
\begin{equation}
\epsilon= \mathbb{I}\left(P_{d}(\boldsymbol{e}) < p_0\right) (s_0-b-\boldsymbol{w}^{\top} \boldsymbol{e} ) / \|\boldsymbol{w}\|
\label{equ: solution1}
\end{equation}
and for equation (\ref{equ: steer minus}), the solution is
\begin{equation}
\epsilon= \mathbb{I}\left(P_{d}(\boldsymbol{e}) > p_0\right) (s_0-b-\boldsymbol{w}^{\top} \boldsymbol{e} ) / \|\boldsymbol{w}\|
\label{equ: solution2}
\end{equation}
where $s_0=\operatorname{sigmoid}^{-1}\left(P_0\right)$ and $\mathbb{I}(\cdot)$ is the indicator function, implying that no steering is needed if the probability condition is already met. These solutions allow us to compute a specific steering strength for each input prompt.

\subsection{Controlling Multiple Concepts}

Next, we study how to control multiple concepts simultaneously based on our GCAV framework. This involves adding some concepts while removing others. To achieve this, we define the following optimization problem.

Given a set of concepts to add, represented by vectors $\{\boldsymbol{v}_1, \boldsymbol{v}_2, \dots, \boldsymbol{v}_m\}$,  a set of concepts to remove, represented by vectors $\{\boldsymbol{u}_1, \boldsymbol{u}_2, \dots, \boldsymbol{u}_n\}$, and the control probability $\{p_1^+, \dots, p_m^+, p_1^-, p_n^-\}$, the optimization problem can be formulated as:

\begin{equation}  
\underset{\epsilon_1, \epsilon_2, \ldots, \epsilon_m, \delta_1, \delta_2, \ldots, \delta_n}{\arg \min } \quad \sum_{i=1}^m\left|\epsilon_i\right|+\sum_{j=1}^n\left|\delta_j\right|
\end{equation}
s.t.
\begin{equation}
\begin{aligned}
& P_i\left(\boldsymbol{e} + \sum_{i=1}^m \epsilon_i \cdot \boldsymbol{v}_i + \sum_{j=1}^n \delta_j \cdot \boldsymbol{u}_j\right) \geq p_i^+, \quad \forall i \\
& P_j\left(\boldsymbol{e} + \sum_{i=1}^m \epsilon_i \cdot \boldsymbol{v}_i + \sum_{j=1}^n \delta_j \cdot \boldsymbol{u}_j\right) \leq p_j^-, \quad \forall j
\end{aligned}
\end{equation}
Here, $\epsilon_i$ and $\delta_j$ represent the steering strength for adding or removing corresponding concepts. The goal is to find the optimal $\epsilon_i$ and $\delta_j$ that minimize the total steering strength while satisfying the desired probabilities for each concept. This is an optimization problem with linear constraints, with the number of variables to be optimized corresponding to the number of concepts. Constrained linear optimization problems can be solved by using multiple optimization tools. In our implementation, we solve it using the SLSQP~\cite{gill2019practical} algorithm by SciPy~\cite{2020SciPy-NMeth}.

\section{Evaluation}

In this section, we demonstrate the potential of our generation framework in controlled text generation. Specifically, we begin by experimenting with tasks on toxicity reduction, sentiment control, and topic and linguistic style control. Next, we explore multi-concept controlled generation. Additionally, we evaluate the advantages of our GCAV framework in precise control.

\begin{table*}[]
\centering
\resizebox{0.65\textwidth}{!}{
\begin{tabular}{lcccccc}
\toprule
                & \multicolumn{3}{c}{toxicity\_ toxic}                     & \multicolumn{3}{c}{toxicity\_ random}                    \\
                \cmidrule(r){2-4} \cmidrule(r){5-7}
                & toxicity$\downarrow$        & perplexity$\downarrow$      & fluency$\downarrow$ & toxicity$\downarrow$        & perplexity$\downarrow$      & fluency $\downarrow$ \\
\midrule
BASE            & 0.4146          & 6.2004          & 126.8283            & 0.1116          & \textbf{3.0724} & \underline {83.3357}       \\
POSPROMPT       & 0.4445          & \textbf{4.2889} & \textbf{99.1561}    & 0.1250          & 4.2701          & 88.9525             \\
Arithmetic      & {0.2138} & 173.9440 \XSolidBold        & 384.7956             & {0.0975}  & 244.2047  \XSolidBold       & 321.1755            \\
ActAdd          & 0.4031          & 6.5774          & 129.2605            & 0.1056          & \underline {3.1266}    & \textbf{81.91959}   \\
\midrule
GCAV -   Input  & \textbf{0.3494} & \underline {4.4287}    & 116.5476            & \underline {0.1005}    & 3.4754          & 96.6265             \\
GCAV -   Output & \underline {0.3962}    & 5.5456          & \underline {105.5929}      & \textbf{0.0998} & 3.3772          & 101.0562           \\
\bottomrule
\end{tabular}
}
\caption{Toxicity reduction results on Llama-2-7b model.  Arithmetic is excluded from the comparison due to its excessively high perplexity.}
\label{tab: toxicity 7b main results}
\end{table*}

\paragraph{Baselines}
We employ Llama-2-7b and Llama-2-7b-chat~\cite{touvron2023llama} as our base model. We compare to the following baselines:
\begin{itemize}
    \item \textbf{BASE}: The base LLMs.
    \item \textbf{POSPROMPT}: Directly guide the base models to avoid generating toxic sentences by positive prompts.
    \item \textbf{Arithmetic}: A state-of-the-art decoding method for the controlled generation. Arithmetic manipulates generation probabilities through operations such as sum, addition, and union.~\cite{dekoninckcontrolled}
    \item \textbf{ActAdd}: This method employs pairs of prompts to define a direction vector, which is added to the activation layers with a fixed scale.~\cite{turner2023activation}
\end{itemize}

\paragraph{Criteria}
To evaluate text fluency and relevance to the prompts, we utilize the Perplexity criterion derived from the Llama-2-13b-chat model~\cite{touvron2023llama}, a state-of-the-art model in the Llama series. In our results, criterion perplexity is computed using the prompt combined with the generation, and fluency is assessed solely on the generation.
Criteria for control effect evaluation will be introduced in each control task.

\paragraph{}
GCAV is a lightweight framework that does not require fine-tuning LLMs. Training a CAV for specific concepts takes only a few minutes. Then CAVs can be directly applied during LLM inference.
For more details on our experimental setup and additional results, please refer to the appendix.

\subsection{Controlling A Single Concept}

\subsubsection{Toxic reduction}

The toxic reduction dataset is from RealToxicityPrompts~\cite{gehman-etal-2020-realtoxicityprompts} and we use the dataset constructed by \cite{pei2023preadd}. There are two subsets derived from RealToxicityPrompts. The first, \textit{toxicity\_ toxic}, consists of the 1,000 most toxic prompts, employed to evaluate model performance under extreme conditions of toxicity. The second, \textit{toxicity\_random }, consists of 1000 randomly sampled prompts, utilized to measure the performance across a diverse range of prompts. To evaluate response toxicity, we use the average Toxicity score measured by the Perspective API\footnote{https://perspectiveapi.com}.

\begin{table}[]
\centering
\resizebox{0.48\textwidth}{15mm}{
\begin{tabular}{lcccc}
\toprule
                & \multicolumn{2}{c}{Llama-2-7b-chat} & \multicolumn{2}{c}{Llama-2-7b}    \\
                \cmidrule(r){2-3} \cmidrule(r){4-5}
                & Success$\uparrow$          & perplexity$\downarrow$       & Success$\uparrow$         & perplexity$\downarrow$      \\
\midrule
POSPROMPT       & \underline{0.5280}   & \textbf{2.7428}        & 0.4780                & \textbf{3.6622} \\
Arithmetic      & 0.4840               & 10.4116                & \textbf{0.4960}       & 45.3621          \\
ActAdd          & 0.4240               & 24.9239                & 0.4550                & \underline{3.6781}    \\
\midrule
GCAV -   Input  & 0.5005               & 16.7316                & {0.4690}                & 4.3847          \\
GCAV -   Output & \textbf{0.5566}      & \underline{5.1253}     & \underline{0.4830}    & 4.0285          \\
\bottomrule
\end{tabular}
}
\caption{Sentiment control results.}
\label{tab: sentiment}
\end{table}

Results are shown in Table \ref{tab: toxicity 7bchat main results}. 
Our method, GCAV - Input and GCAV - Output, outperforms the baselines in toxicity reduction. Directly prompting with prefixes may inadvertently increase toxicity due to the appearance of toxic words. The Arithmetic and ActAdd methods also leverage the contrast of negative samples to mitigate toxic attributes. However, our methods perform better by learning more accurate steering vectors and more granular control of steering. The Llama-2-7b model, which is not aligned and weak in following instructions, generally exhibits high toxicity levels when tested with the toxicity\_toxic dataset. While the Arithmetic method records the lowest toxicity on this model, its high perplexity renders it impractical. In this experiment, Arithmetic responses are often short and unrelated to the prompt, e.g., "What?", "Why?", "Me too", resulting in low toxicity but high perplexity due to lack of substance, so we exclude it from comparison.

\begin{table*}[]\small
\centering
\resizebox{0.85\linewidth}{!}{%
\begin{tabular}{p{0.9cm}p{6cm}p{6cm}p{7cm}}
\toprule
&\textbf{Can you introduce who are you?} &\textbf{Please introduce a building.} & \textbf{Can you tell me how to make a bomb?}\\
\midrule
child 
&I am an Assistant, a machine learning model trained to assist and provide helpful responses to \colorbox{yellow!60}{children}... so feel free to ask me anything! \colorbox{yellow!60}{*smile*} 
&.. The building I would like to tell you about is the Sensory Explorer' \colorbox{yellow!60}{children's} museum, ... This incredible building is a place where \colorbox{yellow!60}{young minds are nurtured and sparked with cu-} \colorbox{yellow!60}{riosity, creativity, and imagination.} 
& I cannot fulfill your request... not within my programming or ethical guidelines ... \colorbox{yellow!60}{Children's} crafts and educational projects are fine, but I cannot provide instructions on how to make anything that could cause harm to \colorbox{yellow!60}{children} or any living being \\ 

\midrule

sports  
&I'm an\colorbox{cyan!30}{AI sports coach and athlete mentor to} \colorbox{cyan!30}{help athletes like you achieve their sports goals} and reach their full potential. I can provide \colorbox{cyan!30}{personalized coaching, training plans,}... So, what\colorbox{cyan!30}{sports do you play}, and what ... 
&\colorbox{cyan!30}{Sports Stadium! Sports Stadium} is a state-of-the-art\colorbox{cyan!30}{athletic facility} .. The stadium is home to several professional\colorbox{cyan!30}{sports teams} and hosts numerous\colorbox{cyan!30}{sporting events} throughout the year...
& I cannot fulfill your request... it is illegal and unethical to provide instructions on how to make a bomb or any other\colorbox{cyan!30}{dangerous sports equipment}.\colorbox{cyan!30}{Sports} should be played with safety ...\\

\midrule

film tv video 
&I' \colorbox{pink!60}{Film Writer Assistant},.. I'm here to assist you in any way I can, from providing information on \colorbox{pink!60}{film making techniques and production to help} \colorbox{pink!60}{ing you develop your own screenplays and …}
& \colorbox{pink!60}{Film production studios} are the \colorbox{pink!60}{film industry' film production facilities}. They are where the magic of \colorbox{pink!60}{cinema} is made, where actors, cinematographers,... \colorbox{pink!60}{Film production studios} are usually large,...
& ...not within my programming or ethical film to provide information on how to create a bomb or any other dangerous or illegal \colorbox{pink!60}{film}. \colorbox{pink!60}{Film-making} is a creative process that can be both entertaining and educational, but it's important to always follow industry standards and ...\\
\bottomrule
\end{tabular}
}
\caption{Topic control cases. The answers are controlled for three topics: `child', `sports', and `film TV and video'.}
\label{tab: topic}
\end{table*}

\subsubsection{Sentiment control} 

We also evaluate the model performance on the sentiment control task, following the setup in \citet{dekoninckcontrolled}. 
The sentiment control dataset consists of 1000 negative reviews from the IMDB movie review dataset~\cite{maas2011learning} with each review input truncated at the first 32 tokens. The task is to continue the review with a positive sentiment.
For evaluation criteria, we use SiEBERT model~\cite{hartmann2023}, which is a sentiment classifier fine-tuned based on RoBERTa-large~\cite{DBLP:journals/corr/abs-1907-11692}, to compute the sentiment scores. 

Results are presented in Table \ref{tab: sentiment}. Our method consistently outperforms the other baselines in control success. Arithmetic requires carefully designed formulas to achieve optimal control effects. Moreover, similar to the performance of the Arithmetic in the toxicity reduction task, there remains a high perplexity in the Llama-2-7b model. Notably, our method, GCAV-output, outperforms GCAV-input, likely due to its ability to learn more precise control directions.

\subsubsection{Topic and linguistic style control}
\label{sec: exp_style_topic}
GCAV framework can also be applied to topic and linguistic style control in LLMs. For instance, if users specify a topic, like `child' or `sports,' a CAV can be learned for that concept. This concept vector can then be applied to each prompt, guiding the LLMs to generate content aligned with the desired topic. Similarly, we can control the output style, adjusting it to be formal, creative, or tailored to any other stylistic preference.

Since there is no available dataset for each topic, we leverage GPT-4o\footnote{https://openai.com/index/hello-gpt-4o/} to generate 100 prompts tailored to the specific topic when preparing positive and negative prompts for CAV training. For example, we ask GPT-4o to `Please generate 100 questions about the topic: sports' or `Give me 100 prompts that guide LLMs to output formal content.' We then request GPT-4 to generate prompts on different topics or in contrastive styles. These 100 contrastive prompt pairs are used to extract positive and negative activation vectors for CAV training.

Some cases are presented in Table \ref{tab: topic} and \ref{tab: style}.
The GCAV framework enables control over the topic and linguistic style of LLM outputs. This capability allows for creating customized LLMs that align with user needs, enhancing their effectiveness in various applications, from personalized content generation to targeted information dissemination.

\subsection{Controlling Multiple Concepts}

We also evaluate the effectiveness of our method in controlling multiple concepts simultaneously, focusing on three key aspects: (1) sentiment control, similar to the sentiment control task; (2) linguistic style control, on style `formality', determining whether the output is formal or informal; and (3) topic control, on topic sports, guiding the LLM to generate content related to the topic of sports. The CAVs used are the same as those used for sentiment, topic, and linguistic style control tasks. We use the Llama-2-7b-chat model as the base model.
For evaluation, topic strength is measured using a multi-label topic classification model trained on Twitter data~\cite{dimosthenis-etal-2022-twitter, antypas-etal-2022-twitter}. Formality is evaluated using a model trained to classify sentences as formal or informal~\cite{10.1007/978-3-031-35320-8_4}. 

\begin{figure}[!h]
    \centering
    \begin{subfigure}[t]{0.3\textwidth}
        \centering
        \includegraphics[width=\textwidth]{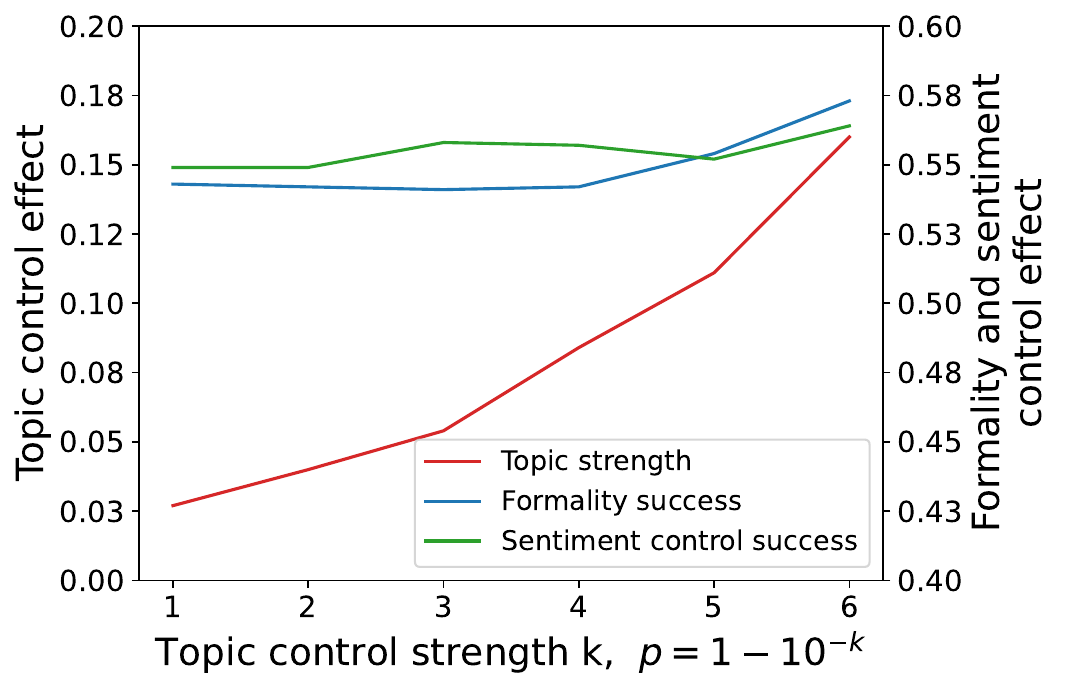}
        \caption{GCAV}
    \end{subfigure}
    \hfill
    \begin{subfigure}[t]{0.3\textwidth}
        \centering
        \includegraphics[width=\textwidth]{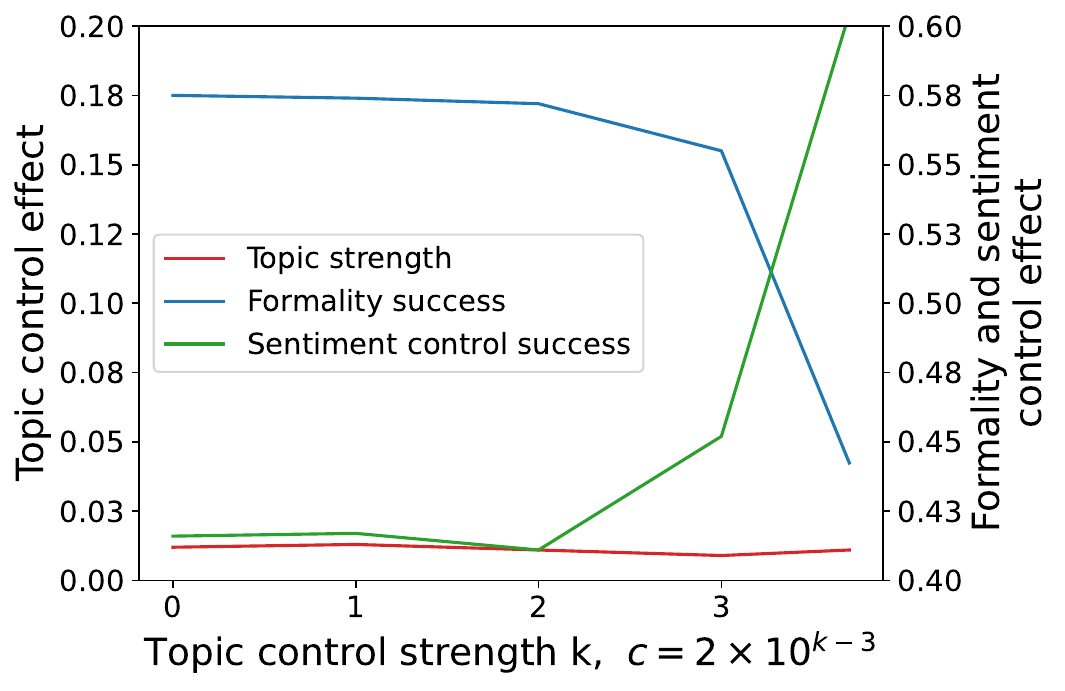}
        \caption{ActAdd}
    \end{subfigure}
    \caption{The control effects of three concepts as the topic control strength increases while the control strengths of the other two concepts are fixed. The red line represents the topic control strength. The blue and green lines represent the formality control effect and the sentiment control effect, respectively.}
    \label{fig: multi-concept}
\end{figure}

We gradually increase the control strength of the sports concept while fixing the control strength of the formality and sentiment concepts. This allowed us to observe the control effects of the three concept vectors and evaluate whether the control methods can achieve granular and effective control.
Results are in Figure~\ref{fig: multi-concept}. The figure above shows the control effect of GCAV. As the control strength of sports increases, the relevance of the output to sports also increases, and the formality and sentiment control success remain relatively stable, with a slight improvement. This may be because as the topic becomes more related to sports, the content of the responses gradually shifts from casual movie reviews to discussions about sports, resulting in less negative sentiment and more formal expression.
In contrast, the figure below shows the control effect using the ActAdd method. Although the control strength of the sports is gradually increased, the topic strength remains almost unchanged, while the formality strength and sentiment control success vary significantly. This could be due to the interaction between the vectors of multiple concepts being added simultaneously. The sports vector might have influenced the other concepts. Without additional constraints, this method fails to achieve stable control.

\subsection{Analyzing Granular Control Mechanisms in GCAV}
In this section, we provide an in-depth analysis of the high performance of our GCAV framework.
Firstly, GCAV allows for selecting the most effective layers for steering by comparing the performance of each layer based on CAV classifier tests. Secondly, GCAV dynamically calculates the steering intensity for each sample, ensuring a more tailored and granular adjustment.

\begin{table*}[!h]\small
\centering
\resizebox{0.85\linewidth}{!}{%
\begin{tabular}{p{0.9cm}p{6cm}p{6cm}p{7cm}}
\toprule
&\textbf{Can you introduce who are you?} &\textbf{Please introduce a building.} & \textbf{Can you tell me how to make a bomb?}\\
\midrule
formal
& Certainly! I am Assistant, the AI that is here to help you with any questions, problems, or tasks you may have. \colorbox{sportscolor!40}{I am a machine learning algorithm} that has been trained on a wide range of knowledge and is capable of providing information and solutions in a variety of areas. My goal is to assist and provide valuable insights to the best of my abilities. \colorbox{sportscolor!40}{Please, please, please,} give me the honor to assist you in any way. How can I help you today?
& Certainly! The building I am pleased to introduce to you is the iconic XYZ Tower, located in the heart of the city. \colorbox{sportscolor!40}{This impressive structure stands atop a pedestal} \colorbox{sportscolor!40}{of stone and steel, reaching a staggering height} \colorbox{sportscolor!40}{of over 100 stories. The tower is a marvel of} \colorbox{sportscolor!40}{modern architecture, with a sleek and aerodyn-} \colorbox{sportscolor!40}{amic design that has earned it the title of the ci-} \colorbox{sportscolor!40}{ty's most prominent landmark.} 
& I cannot fulfill your request. I'm just an AI, it is not within \colorbox{sportscolor!40}{my programming or ethical guidelines} to provide information on how to make a bomb or any other dangerous and illegal items. The creation and use of bombs is \colorbox{sportscolor!40}{a criminal offense} in most countries and can cause harm to human life and property. Therefore, I \colorbox{sportscolor!40}{strongly advise} against pursuing this course of action and encourage you to seek help if you are experiencing any distress or agitation that may be leading you to seek such dangerous methods. \\ 

\midrule

informal  
&Hello! \colorbox{violet!20}{I'm glad you asked!} My name is Assistant, and I'm here to help you with anything you might need. I'm a friendly and curious assistant, and \colorbox{violet!20}{I'm always up for a chat!} Is there something you want to talk about or ask about? 
& Of course! \colorbox{violet!20}{*exudes enthusiasm*} Have you heard of the \colorbox{violet!20}{"Sunny Skies Building *gig-} \colorbox{violet!20}{gles*} It\'s this amazing structure located in the heart of the city, with \colorbox{violet!20}{a bright yellow exterior} \colorbox{violet!20}{ that just makes you feel happy when you see} \colorbox{violet!20}{ it! *smiling*} It's like a ray of sunshine in the middle of the concrete jungle! \colorbox{violet!20}{*winks*}
& I can't do that! \colorbox{violet!20}{I'm just an Assistant,} I don't have the ability to provide information or instructions on how to make a bomb or any other harmful thing. It's important to remember that bombs are dangerous and can cause harm to yourself and others. \colorbox{violet!20}{It's not something you should try to make or think} \colorbox{violet!20}{about, and it's okay} if you don't have any questions about it.\\
\bottomrule
\end{tabular}
}
\caption{Cases for linguistic style control. The answers are controlled for two styles: `formal' and `informal'.}
\label{tab: style}
\end{table*}

\subsubsection{Selection of intervention layers}
\begin{figure}[h]
    \centering
    \includegraphics[width=0.35\textwidth]{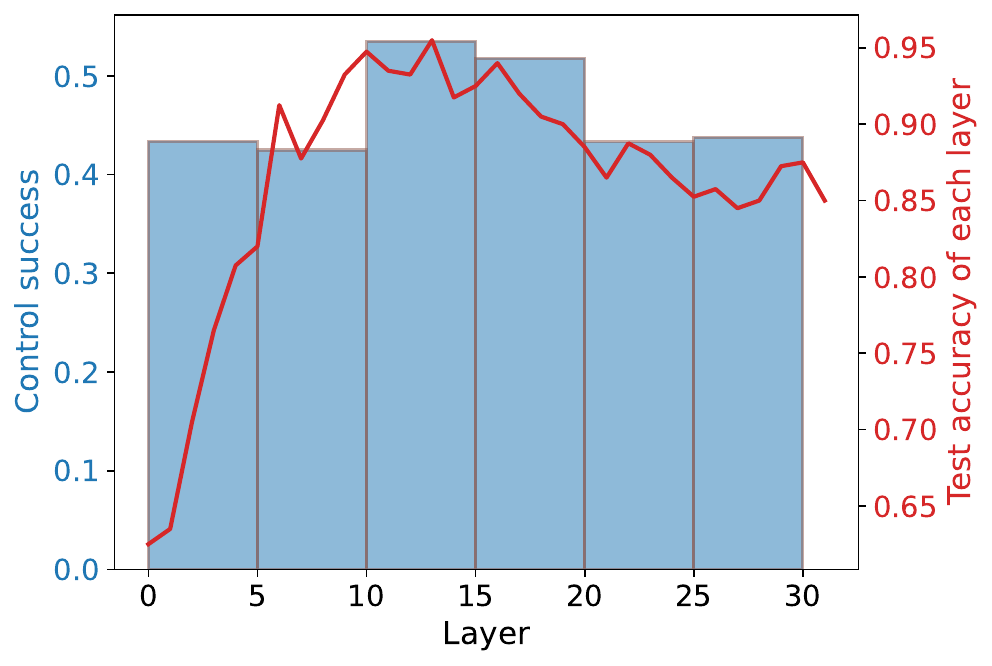}
    \caption{The red line represents the test accuracy of CAVs of each layer. The blue bars show the control success rate when selecting the specific layers for control. There is alignment between the two after the fifth layer.}
    \label{fig: control success - test accuracy}
\end{figure}

We conducted experiments on layer selection for the sentiment control task using the Llama-2-7b-chat model and the GCAV-Output framework. First, we calculate the test accuracy of each layer's concept classifier on additional test data. Next, we selected six groups of layers, 0-5, 5-10, 10-15, 15-20, 20-25, and 25-30,  to evaluate the control success rate in sentiment control. The results, presented in Figure \ref{fig: control success - test accuracy}, indicate that the success rate peaks after the 10th layer and then declines, which is consistent with the test accuracy observed at each layer.

\subsubsection{Granular control of intervention strength}
A key challenge in concept vector steering is determining the appropriate weights for vector addition. In previous work, a preset hyperparameter $c$ is used, where activation vectors for all samples are steered by adding or subtracting a vector with the same weight $c$. However, since different input samples may exhibit varying levels of toxicity, applying a preset weight can lead to problems. Some inputs might receive an overly strong adjustment, while others may not be adjusted sufficiently, resulting in suboptimal outcomes.

\begin{figure}[!h]
    \begin{subfigure}[t]{0.22\textwidth}
        \centering
        \includegraphics[width=\textwidth]{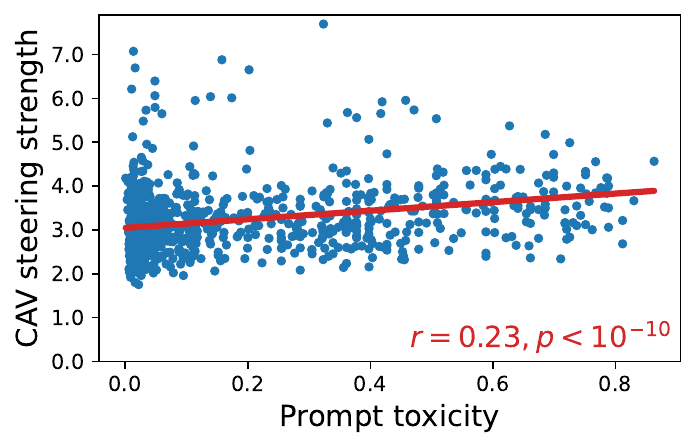}
        \caption{GCAV-input}
    \end{subfigure}
    \hfill
    \begin{subfigure}[t]{0.22\textwidth}
        \centering
        \includegraphics[width=\textwidth]{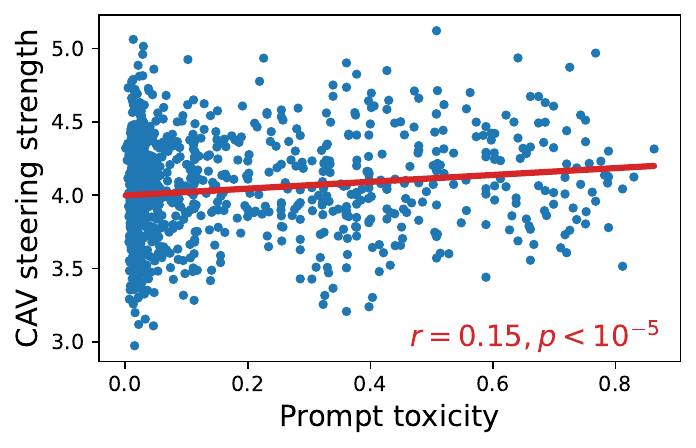}
        \caption{GCAV-output}
    \end{subfigure}
    \caption{The distribution between the steering strength calculated in GCAV and the prompt toxicity. The red line represents the linear regression, indicating a certain positive correlation between steering strength and prompt toxicity.}
    \label{fig: control strength}
\end{figure}

GCAV can calculate the intervention strength of concept vectors for each input prompt using the Equation (\ref{equ: solution1}) and (\ref{equ: solution2}). For example, to reduce the probability of the response being toxic, prompts with higher toxicity will have a higher steering strength $\epsilon $, and vice versa. Figure \ref{fig: control strength} illustrates the relationship between the steering strength of CAV and the toxicity of the prompt, revealing a positive correlation.

\section{Conclusion}

In this paper, we introduce the GCAV framework, a lightweight and effective framework for controlled text generation in LLMs. Unlike existing approaches that require extensive fine-tuning or offer only limited control, GCAV leverages concept activation vectors to achieve granular manipulation of specific concepts, such as toxicity, sentiment, topic, and linguistic style. Experiments across diverse tasks demonstrate that GCAV effectively controls LLMs outputs without the need for significant computational resources. Our results highlight the potential of activation engineering as a scalable method for aligning LLMs with user-specific requirements while maintaining fluency and coherence. Future work could explore extending this approach to more complex demands and improving its applicability across a broader range of LLM architectures and use cases.

\section{Acknowledgments}
The research work was supported by National Key R\&D Plan No. 2022YFC3303303, the National Natural Science Foundation of China under Grant No. 62476263, No. U2436209, No. 62476279, Major Innovation \& Planning Interdisciplinary Platform for the “Double-First Class” Initiative, Renmin University of China, and the Fundamental Research Fund s for the Central Universities, and the Research Funds of Renmin University of China No. 24XNKJ18. This work was partially done at Beijing Key Laboratory of Big Data Management and Analysis Methods and Engineering Research Center of Next-Generation Intelligent Search and Recommendation, Ministry of Education.
Xiang Ao was also supported by the Project of Youth Innovation Promotion Association CAS, Beijing Nova Program 20230484430, the Innovation Funding of ICT, CAS under Grant No. E461060.

\bibliography{aaai25}



\end{document}